\documentclass[letterpaper, 10 pt, conference]{ieeeconf}  

\IEEEoverridecommandlockouts                              
\usepackage{booktabs}  
\usepackage{graphicx}
\usepackage{cuted}
\usepackage{amsmath} 
\usepackage{mdframed}
\usepackage[table]{xcolor}
\usepackage{caption}
\usepackage{float}

\title{\LARGE \bf
Imagine-TAMP: Imagination-Guided Task and Motion Planning in Partial Observability
}

\author{
    Antareep Singha$^{1}$, 
    Shivaram Kumar$^{1}$, 
    Yoonwoo Kim$^{2}$, and 
    Yoonchang Sung$^{1\dagger}$
    \thanks{$^{1}$Nanyang Technological University}
    \thanks{$^{2}$The University of Texas at Austin} 
    \thanks{$^{\dagger}$Corresponding author: {\tt\small yoonchang.sung@ntu.edu.sg}}%
}

\begin{document}

\maketitle
\thispagestyle{empty}
\pagestyle{empty}

\begin{strip}
\centering
\vspace{-4.0em}
\includegraphics[width=\linewidth]{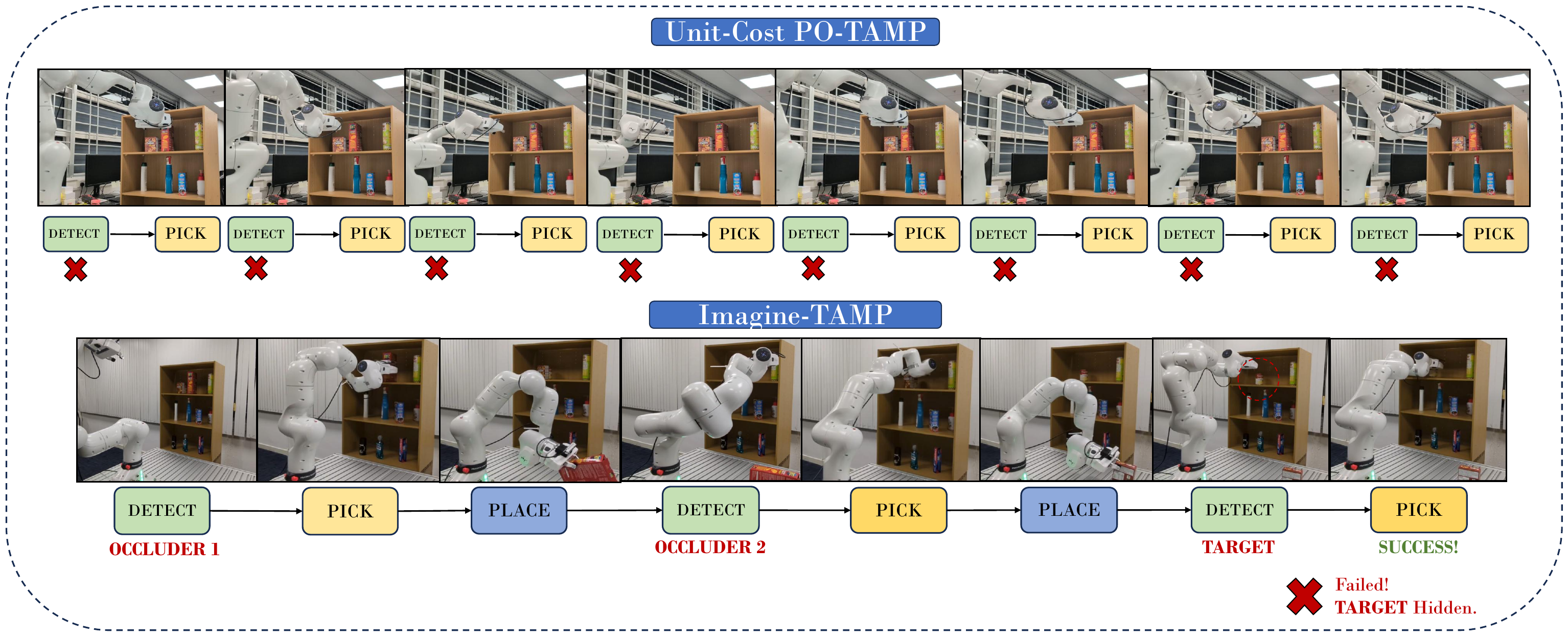}
\captionof{figure}{
Imagine-TAMP \textbf{(bottom)} and Unit-Cost PO-TAMP \textbf{(top)} planners compared in identically arranged shelf environments. The target object, \emph{\textbf{``soup can"}} is occluded within clutter on the top shelf and is not directly visible. Because Unit-Cost PO-TAMP planner does not reason about hypothetical scene geometry or object semantics and favors the shortest plan skeleton, it repeatedly executes uninformative \textsc{Detect} actions until the replanning budget is exhausted. In contrast, \textsc{Imagine-TAMP} leverages imagined scene geometry together with semantic information to identify a more effective plan skeleton: remove the relevant occluding objects and then retrieve the \emph{\textbf{``soup can"}}.}
\label{fig:fig_teaser}
\vspace{-0.25cm}
\end{strip}

\begin{abstract}
Robots operating in cluttered environments must often manipulate objects whose locations are only partially observable.
A central challenge is deciding whether to acquire another observation or to first manipulate objects that may occlude the target.
Conventional task and motion planning (TAMP) approaches typically make this decision using symbolic action costs or expensive geometric planning, neither of which adequately captures how likely an observation is to reveal an occluded target.
We introduce \textsc{Imagine-TAMP}, an interleaved planning and execution framework that uses semantic and geometric imagination to compare alternative task-level strategies under partial observability before committing to expensive motion planning. A vision-language model shapes a particle belief over target locations using commonsense relationships between the target and visible objects, while a generative scene model estimates plausible geometry in unobserved regions. Given a target hypothesis and imagined scene, \textsc{Imagine-TAMP} generates multiple symbolic plan skeletons and assigns non-unit costs that approximate both manipulation effort and target visibility from sensing actions, distinguishing a short but poorly informative observation strategy from a longer strategy that first manipulates an occluder to better expose the target. The selected skeleton is then refined into a feasible continuous plan and executed, with new observations updating the belief and triggering replanning when necessary. Experiments show that imagination-guided evaluation improves observation-versus-manipulation decisions: in viewpoint-constrained shelf scenes, non-unit geometric evaluation increases success from $46.0\%$ to $84.0\%$, while semantic belief shaping further reduces manipulation and replanning. On a real-robot, the complete system reduces planning time by $32\%$ relative to a geometry-only ablation.



\end{abstract}
\vspace{-0.1cm}

\section{Introduction}

Task and Motion Planning (TAMP) provides a principled framework for solving long-horizon robotic manipulation problems by jointly reasoning over discrete task decisions and continuous robot motions \cite{garrett2021integrated,garrett2020pddlstream}. Most TAMP formulations, however, assume that the relevant state of the world is known before planning. This assumption is often violated in realistic environments, where objects may be partially visible, completely hidden by clutter, or located in unobserved regions. A robot searching for an object must therefore reason not only about \emph{how to manipulate in the given environment}, but also about \emph{what information to acquire and when}.

Partially Observable TAMP (PO-TAMP) addresses this problem by maintaining beliefs over uncertain world states and interleaving planning, observation, and execution \cite{garrett2020online,curtis2024partially,kim2026large}. For example, a robot searching for a soup can hidden in a cluttered shelf may move its camera to obtain a better view, remove an object that blocks the target, or directly attempt to reach the target if sufficient evidence suggests that it is accessible. Choosing among these strategies is difficult because their symbolic descriptions alone provide little information about their actual physical movement effort or sensing quality. In many TAMP pipelines, task-level search relies on plan length or simple heuristics, which can favor a shorter plan even when its continuous realization is expensive or its sensing action provides poor visibility. 

Consider the example in Fig.~\ref{fig:fig_teaser}. Suppose that a target object is believed to lie behind one of several visible objects. Under unit action costs, a symbolic planner may prefer a plan skeleton that directly performs \textsc{Detect} because it contains fewer symbolic actions than an alternative that first picks and places an occluding object before detecting the target. However, the selected camera viewpoint may provide little additional visibility of the target, whereas removing an occluder may expose it far more effectively despite requiring additional actions. Determining which strategy is preferable therefore requires reasoning beyond symbolic plan length, accounting for both the \emph{visibility provided by candidate observations} and the \emph{physical effort required to realize them}.

Recent generative perception models provide a new opportunity for addressing this problem. Rather than treating unobserved regions as simply unknown, these models can generate plausible hypotheses about hidden scene geometry. At the same time, vision-language models (VLMs) encode semantic relationships that provide useful commonsense priors about where an object is likely to be. For instance, when searching for a soup can, a VLM may assign greater probability to regions behind other food containers than to regions behind unrelated objects. These two forms of reasoning provide complementary information: semantic reasoning estimates \emph{where the target is likely to be}, while geometric imagination provides a concrete scene hypothesis from which the robot can evaluate \emph{how to observe or expose the target}.

We introduce \textsc{Imagine-TAMP}, an interleaved PO-TAMP framework that uses semantic and geometric imagination to guide high-level task decisions. The framework maintains a belief over possible target poses and uses VLM-derived semantic information to estimate where the target is likely to be. Given a target hypothesis and imagined scene geometry, \textsc{Imagine-TAMP} generates multiple candidate symbolic plan skeletons and cheaply estimates the geometric cost of realizing each strategy before full motion planning. Manipulation actions are evaluated using approximate motion costs, while \textsc{Detect} actions additionally account for how well candidate camera viewpoints expose the hypothesized target. Propagating these action-level estimates across a skeleton, lets the planner directly compare qualitatively different strategies, such as observing immediately or first manipulating an occluder to obtain a better observation.


The most promising skeleton is passed to the underlying TAMP solver for full continuous refinement and subsequent execution. Candidate continuous parameters, such as camera viewpoints, are ranked by a lightweight evaluator and used to guide this refinement. After new observations or scene changes, the belief and scene representation are updated, and the process repeats. The robot therefore repeatedly \emph{believes, imagines, evaluates, plans, and acts}.

Our main contributions are:
\begin{itemize}
\item We introduce an imagination-guided PO-TAMP framework that integrates semantic target-location reasoning and geometric scene hypotheses to choose between information-gathering and scene-manipulation actions.
\item We propose a non-unit-cost task-plan evaluator that combines approximate manipulation costs with imagination-based sensing costs, enabling qualitatively different plan skeletons to be compared before expensive continuous refinement.
\item We evaluate the framework on partially observable manipulation tasks with varying target locations, clutter, and occlusion, demonstrating improved task efficiency relative to unit-cost and ablated planning baselines.
\end{itemize}


\section{Related Work}

\textsc{Imagine-TAMP} builds on work in partially observable TAMP and object search, generative reasoning about unobserved geometry, and foundation-model-guided robot planning. We focus on how these approaches incorporate uncertainty, geometric imagination, and semantic knowledge into planning, and distinguish our use of imagined geometry for evaluating alternative sensing and manipulation strategies.

\subsection{Task and Motion Planning under Partial Observability}

TAMP integrates symbolic task planning with continuous geometric reasoning for long-horizon manipulation~\cite{garrett2021integrated,garrett2020pddlstream}. Early work extended this paradigm to belief space by jointly reasoning about task actions, perception, and state estimation~\cite{kaelbling2013integrated,hadfield2015modular}. SS-Replan~\cite{garrett2020online} instead uses determinization and online replanning, while TAMPURA~\cite{curtis2024partially} explicitly reasons about uncertainty, information gathering, and risk. More recently, PO-LGP~\cite{phiquepal2026optimizing} reasons over decision trees at the task level and trajectory trees at the motion level. Related active-perception methods optimize robot motion to acquire task-relevant visual information~\cite{jauhri2024active}, while object-search methods reason about manipulating occluders or changing viewpoints to reveal hidden targets~\cite{wong2013manipulation,dogar2014objectsearch,lin2015occludedsearch}. Recent TAMP methods also use anticipatory costs to distinguish among feasible plans~\cite{dhakal2026anticipatory}. In contrast, \textsc{Imagine-TAMP} compares alternative sensing--manipulation plan skeletons using lightweight estimates of physical effort and target visibility before continuous refinement.

\subsection{Reasoning about Unobserved Geometry}

Amodal reconstruction and scene-completion methods infer geometry hidden from the current observation for downstream robotic manipulation~\cite{agnew2021amodal,agarwal2026scenecomplete,wu2025amodal3r}. Generative models have also been used to infer unobserved objects and scene structure as priors for robot planning~\cite{bhattacharjee2025uod,bhattacharjee2025into}, and more broadly to imagine desired or future states for manipulation~\cite{huang2025imaginationpolicy,barcellona2025dream}. Our objective differs from accurate scene reconstruction or direct action generation: imagined geometry serves as a planning heuristic for estimating otherwise unobservable consequences of candidate actions, such as how effectively moving an occluder may expose a hypothesized target. Final motions are subsequently generated and verified by the underlying model-based TAMP system.

\subsection{Foundation Models for Semantic Planning Guidance}

Language and vision-language models have been integrated with TAMP to provide commonsense guidance while retaining geometric reasoning. LLM-GROP uses language-model knowledge of plausible object arrangements to guide TAMP~\cite{ding2023task}, while VLM-TAMP uses VLM-generated intermediate subgoals to guide long-horizon TAMP~\cite{yang2025guiding}. OWL-TAMP uses VLM-generated discrete and continuous constraints for open-world manipulation~\cite{kumar2026owltamp}, and related methods use foundation models to guide task and motion choices or recovery~\cite{kwon2025kinodynamic}. Most closely related is \textsc{CoCo-TAMP}~\cite{kim2026large}, which uses commonsense knowledge to shape beliefs over unseen object locations during interleaved planning and execution. We similarly use semantic reasoning to bias the target belief, but combine it with geometric imagination: semantic reasoning estimates \emph{where} the target is likely to be, while imagined geometry estimates \emph{how} alternative sensing or manipulation actions may expose it. The VLM therefore shapes the belief rather than directly serving as the planner or geometric verifier.

\begin{figure*}[t] 
\centering 
\includegraphics[width=\linewidth]{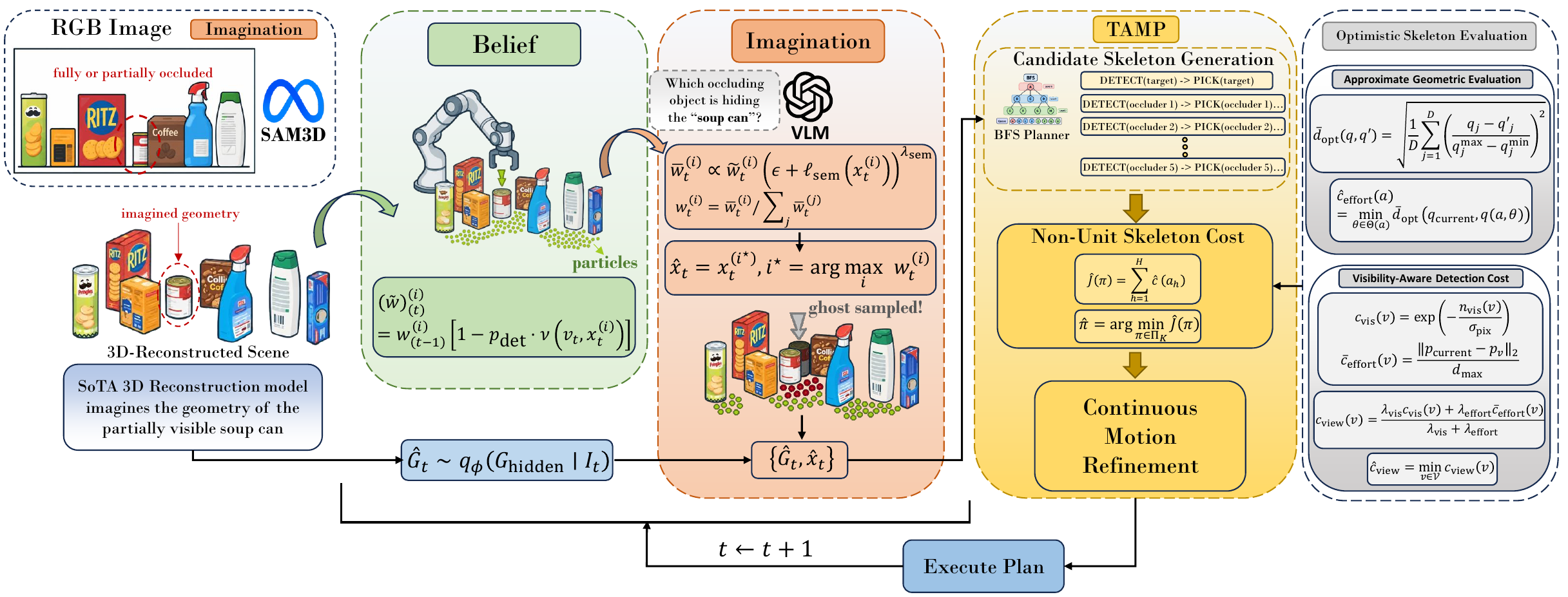} 
\caption{
\textsc {Imagine-TAMP} is an imagination-driven interleaved planning and execution framework for long horizon tasks under partial observability. It uses a 3D scene-completion model, SAM~3D, to reconstruct the scene. The reconstructed scene is spawned in a simulator, with particle beliefs over hidden object locations initialized uniformly and updated through observations. A VLM provides semantic scores to occluding objects based on their likelihood of hiding the target. Candidate plan skeletons are then evaluated using non-unit, optimistic heuristic costs that account for factors such as visibility, motion effort, and geometric feasibility. The lowest-cost skeleton is subsequently refined into a feasible continuous plan and executed in the real world.}
\label{fig:fig_pipeline} 
\vspace{-0.2cm}
\end{figure*}

\section{Problem Formulation}
\label{sec:prob}

We consider a robot operating in a cluttered workspace containing a set of observed objects $\mathcal{O}_t = \{o_1,\ldots,o_{|\mathcal{O}_t|}\}$ and a task-relevant target object $o^\star$ whose pose may be partially or completely unobserved. The target is part of a broader manipulation task: the robot must acquire sufficient information about its location while performing manipulation actions needed to accomplish the task goal. Let $x^\star \in \mathcal{X}$ denote the unknown target pose and $b_t(x^\star)$ denote the robot's belief over this pose at planning step $t$. For clarity, we formulate the method for a single target under partial observability, while the framework can be extended to multiple targets.

The robot can execute symbolic manipulation and sensing actions such as $a \in \{\textsc{Pick}(o), \textsc{Place}(o,r), \textsc{Detect}(o)\},$ where $r$ denotes a placement region. Each symbolic action is initially specified only at the task level and must later be instantiated with continuous parameters, such as a grasp pose, placement pose, or camera viewpoint, together with a collision-free robot trajectory.

A symbolic plan skeleton is $\pi=(a_1,\ldots,a_H),$ where the continuous parameters of each action are initially unspecified. Our objective is to select and execute actions that accomplish the manipulation task while minimizing cumulative physical and sensing cost, $J(\pi)=\sum_{h=1}^{H}c(a_h),$ under uncertainty about $x^\star$ and the geometry of currently unobserved space.

Solving the complete belief-space TAMP problem directly is computationally expensive because it requires reasoning over future observations, belief updates, and continuous geometric choices. Following a determinize-and-replan strategy~\cite{garrett2020online,kim2026large}, we instead instantiate a deterministic planning problem from the current belief, solve it using a conventional TAMP planner, and replan whenever execution provides new information. Within each planning iteration, we seek an inexpensive estimate $\widehat{J}(\pi)$ that can rank a collection of candidate skeletons before continuous motion refinement.


\section{Imagine-TAMP}
\label{sec:method}

Fig.~\ref{fig:fig_pipeline} summarizes \textsc{Imagine-TAMP}. Rather than solving the full belief-space TAMP problem, we adopt a determinize-and-replan strategy. At each planning iteration, the framework (1) updates and semantically shapes the target belief, (2) constructs a deterministic scene hypothesis using imagined hidden geometry and a hypothesized target pose, (3) generates and approximately evaluates multiple candidate TAMP skeletons, and (4) refines and executes the most promising skeleton using the underlying TAMP solver. New observations update the belief and trigger another planning iteration until the target is found and the task is completed.

\subsection{Particle Belief over the Hidden Target}
\label{sec:belief}

We represent the belief $b_t$ over the target pose using a weighted particle set $B_t=\{(x_t^{(i)},w_t^{(i)})\}_{i=1}^{N}$, where $x_t^{(i)}$ denotes a candidate target pose, $w_t^{(i)}$ its probability, and $\sum_i w_t^{(i)}=1$. When no object-specific prior is available, particles are initialized over geometrically feasible supporting surfaces or volumes.

Let $z_t$ denote the observation obtained from camera viewpoint $v_t$, and let $\nu(v_t,x_t^{(i)})\in[0,1]$ denote the estimated visibility of particle $x_t^{(i)}$ from that viewpoint. When the target is not detected, particles corresponding to locations that should have been visible are downweighted according to $\widetilde{w}_{t}^{(i)}=w_{t-1}^{(i)}[1-p_{\mathrm{det}}\nu(v_t,x_t^{(i)})]$, where $p_{\mathrm{det}}$ is the nominal probability of detecting a visible target. The weights are subsequently reshaped by the semantic prior of Sec. \ref{subsec:vlm_update}, normalized, and resampled when necessary, causing the belief to concentrate progressively on regions that remain consistent with the observations.

\subsection{VLM-Based Semantic Belief Shaping}
\label{subsec:vlm_update}
Geometric observations alone do not distinguish between hiding regions that are equally feasible geometrically but differ in plausibility. We therefore query a VLM with the current RGB observation $I_t$ and a description of the search problem: the target $o^\star$ and the candidate occluders(by name if object identities are available, or by colour and shape), and ask it to identify the visible objects most likely to conceal the target. It is asked for a discrete judgment, which we convert to a semantic score $s_j\in[0,1]$ for
each visible object $o_j\in\mathcal{O}_t$. When the answer is a ranked list, the $k$-th ranked object receives $s_j\in\{1.0,\,0.1,\,0.05,\,0.02\}$ for $k\le4$ and $0.01$ thereafter, with unranked objects at $0$; when the answer
names a group of objects(cluster), every member receives $s_j=1$ and all others $0$.

Each particle is associated with an occluder according to the occlusion region in which it lies. From the camera position $\mathbf{c}$, occluder $o_j$ with center $\mathbf{o}j$ and bounding radius $r_j$ defines an approximate conical occlusion region with half-angle $\alpha_j=\arcsin\left(r_j/\lVert\mathbf{o}j - \mathbf{c}\rVert\right)$; a particle lies in the occlusion region of $o_j$ if it falls within this cone and lies behind the front surface of $o_j$ relative to the camera. Let $\rho(x_t^{(i)})$ denote the occluder with the largest VLM semantic score $s_j$ among those whose occlusion regions contain $x_t^{(i)}$, and let $\ell_{\mathrm{sem}}(x_t^{(i)})=s_{\rho(x_t^{(i)})}$, or $0$ if the particle lies in no occlusion region. After the observation update, the weights are set to $\bar{w}_t^{(i)}\propto\widetilde{w}_t^{(i)}\big(\epsilon+\ell_{\mathrm{sem}}(x_t^{(i)})\big)^{\lambda_{\mathrm{sem}}}$, $w_t^{(i)}=\bar{w}_t^{(i)}\Big/\sum\nolimits_j\bar{w}_t^{(j)}$, where $\lambda_{\mathrm{sem}}$ is the semantic guidance strength and $\epsilon=10^{-4}$ ensures that every geometrically consistent particle retains nonzero weight. The VLM prediction therefore acts as a soft prior on search order: an incorrect preference delays the search rather than excluding the true location, and is corrected by subsequent negative observations through the update of Sec.~\ref{sec:belief}.

\subsection{Imagining Hidden Geometry}

Evaluating candidate plans is difficult because their utility often depends on geometry that is currently occluded. Let $G_{\mathrm{hidden}}$ denote the geometry of the currently unobserved portion of the workspace. We use a generative scene model $p_{\phi}$ to produce a plausible geometric hypothesis conditioned on the current visual observation, $\widehat{G}_t \sim p_{\phi}(G_{\mathrm{hidden}} \mid I_t)$, where $I_t$ denotes the current RGB observation.

The imagined geometry is not treated as ground truth or used for execution; it is a lightweight hypothesis for evaluating otherwise unobservable quantities, such as whether a candidate viewpoint would expose the hypothesized target. Final motions are still generated and checked by the underlying TAMP and collision-checking pipeline.

\begin{figure}[ht]
    \centering
    \includegraphics[width=1.0\linewidth]{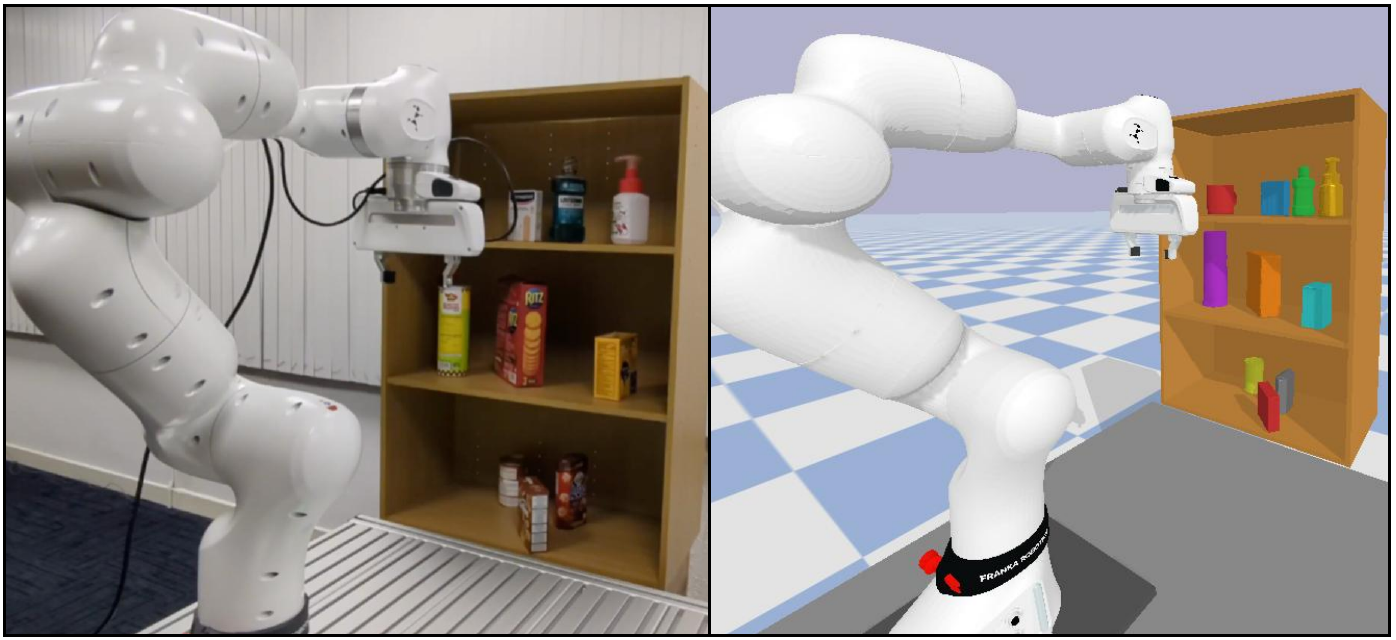}
        \caption{Real-world shelf scene(left) and the realized scene on PyBullet with SAM3D reconstructed occluding objects(right). Imagine-TAMP hypothesizes on the realized scene before proceeding with real-world execution.}
    \label{fig:3d_model}
    \vspace{-0.4cm}
\end{figure}

\subsection{Belief-Guided Determinization}

To obtain a deterministic planning problem from the target belief, we select a representative target hypothesis from the semantically shaped belief. In our implementation, we use the maximum-weight particle $\hat{x}_t = x_t^{(i^\star)}, i^\star=\arg\max_i\, w_t^{(i)}$, and insert a temporary \emph{ghost target} at $\hat{x}_t$ into the planning scene.

The ghost target does not assert that the object is actually located at $\hat{x}_t$. Rather, together with the imagined scene geometry, it provides a concrete deterministic scene from which the task planner can generate candidate symbolic strategies and continuous parameters. Whenever observations invalidate the current hypothesis, the belief is updated and a new deterministic planning problem is constructed.

\subsection{Candidate Skeleton Generation}

Given the current symbolic state and determinized scene hypothesis, we construct a symbolic search tree using breadth-first search (BFS). Each node represents a symbolic state and each edge corresponds to an applicable high-level action. BFS expands action sequences in increasing skeleton depth until up to $K$ goal-reaching plan skeletons have been collected, $\Pi_K=\{\pi_1,\ldots,\pi_K\}$. At this stage, no continuous refinement is performed: the skeletons specify only the high-level action sequence.

For example, two candidate strategies may be $\pi_1:\textsc{Detect}(o^\star)\rightarrow\textsc{Pick}(o^\star)$ and $\pi_2:\textsc{Detect}(o_a)\rightarrow\textsc{Pick}$ $(o_a)\rightarrow\textsc{Place}(o_a,r)\rightarrow\textsc{Detect}(o^\star)\rightarrow\textsc{Pick}(o^\star)$. Because BFS enumerates shallower skeletons first, selecting plans by symbolic depth would alone favor $\pi_1$, even when its detection action leaves the target poorly visible. We therefore re-rank the skeletons using lightweight geometric costs that approximate both manipulation effort and the visibility quality of sensing actions before invoking full motion planning.

\subsection{Approximate Geometric Evaluation}

Motion-level refinement of every skeleton in $\Pi_K$ would be computationally expensive. We instead instantiate a small set of candidate continuous parameters for each action and use inexpensive geometric surrogates to estimate their cost.

For a manipulation action $a\in\{\textsc{Pick},\textsc{Place}\}$, let $\Theta(a)$ denote a set of sampled continuous parameters, such as grasp poses or placement poses, and let $q(a,\theta)$ denote the robot configuration associated with action $a$ instantiated using parameter $\theta$. Each sampled parameter $\theta\in\Theta(a)$ is retained only if it admits an inverse-kinematics solution that is feasible in the determinized scene. If none exist, the action is deemed infeasible and assigned a high penalty. To obtain a dimensionless \emph{effort} cost, i.e, a measure of the arm motion required to realize an action; we normalize each joint displacement by its valid joint range and define $\bar{d}_{\mathrm{opt}}(q,q')=\sqrt{\frac{1}{D}\sum_{j=1}^{D}\left(\frac{q_j-q'_j}{q_j^{\max}-q_j^{\min}}\right)^2}$, where $D$ is the number of robot joints and $q_j^{\min}$ and $q_j^{\max}$ denote the joint limits. We then estimate the manipulation effort cost as $\hat{c}_{\mathrm{effort}}(a)=\min_{\theta\in\Theta(a)}\bar{d}_{\mathrm{opt}}(q_{\mathrm{current}},q(a,\theta))$. This lightweight estimate is used only to rank candidate skeletons and does not replace full collision-free motion planning.

\subsection{Visibility-Aware Detection Cost}
\label{sec:detection-cost}

A \textsc{Detect} action must account not only for the motion required to reach a camera viewpoint but also for whether that viewpoint actually exposes the hypothesized target. We therefore generate a set of candidate viewpoints $\mathcal{V}=\{v_1,\ldots,v_M\}$ around the ghost target, using a hemisphere of camera poses for tabletop scenes and a frontal grid of camera poses for shelf scenes.

For each candidate viewpoint $v$, we render a low-resolution $64\times64$ segmentation image of the imagined scene toward the hypothesized target. Let $n_{\mathrm{vis}}(v)$ denote the number of rendered pixels labeled as the target object. We convert this quantity into a visibility cost $c_{\mathrm{vis}}(v)=\exp(-n_{\mathrm{vis}}(v)/\sigma_{\mathrm{pix}})$, where $\sigma_{\mathrm{pix}}$ is a scaling parameter. A fully occluded target therefore yields a cost near one, and greater exposure yields progressively smaller costs.

We additionally estimate the effort required to reach viewpoint $v$, again as arm motion, using the Euclidean distance between the current arm position and the candidate viewpoint. To place this cost on the same dimensionless scale, we normalize it by a fixed workspace distance $d_{\max}$ and define $\bar{c}_{\mathrm{effort}}(v)=\lVert \mathbf{p}_{\mathrm{current}}-\mathbf{p}_v\rVert_2/d_{\max}$, where $\mathbf{p}_{\mathrm{current}}$ denotes the current arm position and $\mathbf{p}_v$ the candidate viewpoint position. The resulting viewpoint score combines the two cost terms used throughout, visibility and effort, as $c_{\mathrm{view}}(v)=(\lambda_{\mathrm{vis}}c_{\mathrm{vis}}(v)+\lambda_{\mathrm{effort}}\bar{c}_{\mathrm{effort}}(v))/(\lambda_{\mathrm{vis}}+\lambda_{\mathrm{effort}})$, where the weights $\lambda_{\mathrm{vis}}$ and $\lambda_{\mathrm{effort}}$ balance target visibility against arm-motion effort.

The approximate cost of the \textsc{Detect} action is given by the lowest-cost candidate viewpoint, $\hat{c}_{\mathrm{view}}=\min_{v\in\mathcal{V}}c_{\mathrm{view}}(v)$. Importantly, the remaining viewpoints are not discarded. All candidate viewpoints are sorted according to $c_{\mathrm{view}}(v)$ and passed to the underlying TAMP solver in best-first order, allowing the lightweight evaluator to guide continuous planning without assuming that its highest-ranked viewpoint is necessarily motion feasible.

\subsection{Non-Unit Skeleton Cost}
\label{sec:skeleton-cost}

Candidate skeletons are evaluated , with the robot robot and object states propagated according to the action outcome before subsequent actions are evaluated. Consequently, for a skeleton that first moves an occluding object and then executes \textsc{Detect}, the detection cost is evaluated in the resulting hypothetical scene rather than in the initial scene. For a candidate skeleton $\pi=(a_1,\ldots,a_H)$, we compute $\widehat{J}(\pi)=\sum_{h=1}^{H}\hat{c}(a_h)$, where $\hat{c}(a)=\hat{c}_{\mathrm{effort}}(a)$ for $a\in\{\textsc{Pick},\textsc{Place}\}$ and $\hat{c}(a)=\hat{c}_{\mathrm{view}}$ for $a=\textsc{Detect}$.

Unlike unit-cost task planning, this score compares skeletons with different numbers and types of actions according to their approximate geometric realization. So, a short skeleton with a poorly positioned \textsc{Detect} can cost more than a longer one that first exposes the target.

We select $\hat{\pi}=\arg\min_{\pi\in\Pi_K}\widehat{J}(\pi)$ and perform continuous motion refinement only for this skeleton. The score need not predict execution cost accurately; it need only distinguish promising strategies cheaply.

\subsection{Continuous Motion Refinement and Interleaved Execution}

The task-level planner generates alternative symbolic skeletons without performing continuous motion refinement. After approximate evaluation, only the selected skeleton $\hat{\pi}$ is passed to the motion-level refinement stage, which computes feasible grasps, placements, viewpoints, and collision-free trajectories. For a \textsc{Detect} action, the visibility-aware evaluator provides candidate viewpoints in best-first order, and the motion planner searches for a feasible realization.

The refined plan is executed until the task is completed or new information requires replanning. If a \textsc{Detect} action finds the target, the hypothesized target pose is replaced by the observed one; if detection fails, hypotheses that should have been visible from the executed viewpoint are downweighted. The next iteration then repeat belief update, geometric imagination, task-plan evaluation, and continuous refinement on the updated scene.

\section{Experiments}
\label{sec:exp}

We evaluate \textsc{Imagine-TAMP} on partially observable object-search and manipulation tasks in cluttered tabletop and shelf scenes. We investigate three hypotheses:
\begin{enumerate}
\item[\textbf{H1.}] \emph{Non-unit cost evaluation improves the decision between sensing and manipulating the environment, particularly when sensing is constrained or clutter is severe.
}
\item[\textbf{H2.}] \emph{Manipulation enables recovery from severe occlusion, while predicting the visibility consequence of manipulation improves the efficiency of that interaction.}
\item[\textbf{H3.}] \emph{VLM-based belief shaping complements geometric reasoning by directing search toward more plausible target regions and reducing unnecessary manipulation and replanning.}
\end{enumerate}
We first test H1 and H2 without semantic guidance to isolate the contribution of non-unit geometric evaluation, and then evaluate H3 in controlled environments where semantic guidance can be measured directly. Finally, we validate the complete pipeline on a real robot with real perception.

All simulation experiments are conducted in PyBullet using procedurally generated cluttered tabletop and shelf scenes. We use \textsc{cuTAMP}~\cite{shen2025cutamp} for motion-level refinement. Tabletop sensing poses are sampled over a hemisphere around the hypothesized target, whereas shelf sensing is restricted to a frontal grid, creating substantially stronger occlusion constraints. All simulation experiments were run on a system with a dedicated NVIDIA RTX 4060 GPU.

\paragraph{\textbf{Baselines and protocol}}
We compare the following methods:
\begin{itemize}
\item \textbf{Unit-Cost PO-TAMP (UC):} ranks skeletons by the number of symbolic actions.
\item \textbf{Effort-Only:} uses non-unit manipulation costs but disables visibility prediction by setting $\lambda_{\mathrm{vis}}=0$.
\item \textbf{Motion-Cost (MC):} combines manipulation effort with the visibility cost of Sec.~\ref{sec:detection-cost}.
\item \textbf{MC + VLM:} augments MC with VLM-based belief shaping.
\item \textbf{UC + VLM:} combines semantic belief shaping with unit-cost task planning and is the baseline most closely aligned with the semantic belief-shaping mechanism of CoCo-TAMP~\cite{kim2026large}.
\end{itemize}
Effort-Only and MC use the same cost function and differ only in whether the visibility term is active: Effort-Only retains only the effort (arm-motion) term, whereas MC uses both effort and visibility.

Methods are evaluated on identically seeded paired scenes with the same initial belief and occluder configuration. A trial succeeds when the target is detected within 20 replanning steps; detection requires the target to lie within the $60^\circ$ field of view of the selected sensing pose and to be unoccluded along the corresponding line of sight. We report task success, replans, removed occluders, trajectory length, and task-level decision time(\textbf{\emph{t}}) where applicable. Trajectory length is the cumulative planar distance traveled to sensing poses and manipulated occluders; decision time excludes physical execution time. For paired success outcomes, we report $p$-values.

\subsection{H1: When Does Non-Unit Evaluation Matter?}
\label{sec:sim_prelim}

We first compare UC and MC without VLM guidance. Table~\ref{tab:sim_shelf_tabletop} reports $n=50$ paired trials per environment on scenes containing 9--15 randomly placed occluders. On the shelf, where viewpoints are restricted to a frontal grid, MC increases success from $46.0\%$ to $84.0\%$ and reduces average replans from $11.50$ to $4.50$. The methods disagree on 19 scenes, all favoring MC, indicating a significant advantage for MC ($p=3.8\times10^{-6}$). On the tabletop, where hemispherical sampling provides many alternative viewpoints, both methods are nearly saturated ($96.0\%$ for UC and $100.0\%$ for MC, $p=0.5$), although MC still reduces replans from $3.36$ to $2.30$. Thus, non-unit evaluation matters most when viewpoints are restricted: an uninformative observation cannot simply be replaced by another viewpoint, so the planner must instead identify which occluders are feasible and worthwhile to remove.

\begin{table}[htbp]
\centering
\renewcommand{\arraystretch}{1.2}
\setlength{\tabcolsep}{4pt}
\resizebox{\columnwidth}{!}{%
\begin{tabular}{|l|c|c|c|c|c|}
\hline
\textbf{Method} & \textbf{Succ. \%} $\uparrow$ & \textbf{Traj.} & \textbf{Replans} $\downarrow$ & \textbf{Occl.} & \textbf{\emph{t}} $\downarrow$ \\
& [95\% CI] & Len. & & Rem. & (s) \\
\hline
\multicolumn{6}{|l|}{\emph{\textbf{Shelf (frontal-only viewpoints)}}} \\
\hline
Unit-Cost & 46.0 [33,60] & 1.01 & 11.50 & 0.00 & 0.264 \\
\hline
Effort-Only & \textbf{84.0 [71,92]} & 2.63 & 4.84 & 2.46 & 0.151 \\
\hline
Motion-Cost & \textbf{84.0 [71,92]} & 2.37 & \textbf{4.50} & 2.12 & \textbf{0.141} \\
\hline
\multicolumn{6}{|l|}{\emph{\textbf{Tabletop (hemispherical viewpoints)}}} \\
\hline
Unit-Cost & 96.0 [87,99] & 1.07 & 3.36 & 0.00 & 0.211 \\
\hline
Effort-Only & \textbf{100.0 [93,100]} & 1.70 & 2.74 & 1.74 & \textbf{0.191} \\
\hline
Motion-Cost & \textbf{100.0 [93,100]} & 1.34 & \textbf{2.30} & 1.30 & \textbf{0.171} \\
\hline
\end{tabular}%
}
\caption{Unit vs.\ Non-Unit costing on densely-cluttered scenes ($n=50$ paired trials per environment; 95\% confidence interval (CI) in brackets). Trajectory Length and Occluders Removed are reported for diagnosis. Unit-Cost attains the lowest values because it never manipulates(also why it fails). \emph{t} is symbolic execution time in simulation. Effort-Only is included for reference(analyzed in Sec.~\ref{sec:exp-decompose}).}
\label{tab:sim_shelf_tabletop}
\vspace{-0.2cm}
\end{table}

We next vary shelf clutter using $n=20$ paired trials at each occluder count. As shown in Table~\ref{tab:sim_occluder_sweep}, UC degrades from $80.0\%$ success with 5 occluders to $35.0\%$ with 15, whereas MC remains between $85.0\%$ and $90.0\%$. The difference is significant at every level above the sparsest condition. Replans show the same divergence: UC increases from $4.95$ to $14.35$, approaching the 20-replan budget, while MC increases only from $3.10$ to $4.75$. This behavior is consistent with UC repeatedly selecting short \textsc{Detect}-first plans in scenes where additional frontal observations cannot resolve the occlusion. The advantage of non-unit geometric evaluation therefore grows as sensing alone becomes less effective.

\begin{table}[htbp]
\centering
\renewcommand{\arraystretch}{1.2}
\setlength{\tabcolsep}{4pt}
\begin{tabular}{|c|cc|cc|cc|c|}
\hline
\textbf{$N_{\mathrm{occ}}$} & \multicolumn{2}{c|}{\textbf{Success \%} $\uparrow$} & \multicolumn{2}{c|}{\textbf{Traj. Len.(m)}} & \multicolumn{2}{c|}{\textbf{Replans} $\downarrow$} & \textbf{$p$} \\
\cline{2-7}
& \textbf{UC} & \textbf{MC} & \textbf{UC} & \textbf{MC} & \textbf{UC} & \textbf{MC} & \\
\hline
5 & 80.0 & 90.0 & 0.97 & 1.41 & 4.95 & 3.10 & 0.500 \\
\hline
8 & 50.0 & 90.0 & 0.97 & 1.75 & 10.60 & 3.45 & \textbf{0.0078} \\
\hline
12 & 45.0 & 85.0 & 1.02 & 2.28 & 11.90 & 4.40 & \textbf{0.0078} \\
\hline
15 & 35.0 & 85.0 & 1.06 & 2.65 & 14.35 & 4.75 & \textbf{0.0020} \\
\hline
\end{tabular}%
\caption{Shelf occluder-count sweep ($n=20$ paired trials per level). $N_{\mathrm{occ}}$ is the occluder count placed. $p$ denotes the $p$-value for paired task-success outcomes.}
\label{tab:sim_occluder_sweep}
\vspace{-5pt}
\end{table}

\subsection{H2: What Does Visibility Prediction Contribute?}
\label{sec:exp-decompose}

The comparison above mixes two effects: allowing manipulation and predicting which manipulation most improves target visibility. We separate them by sweeping $\lambda_{\mathrm{vis}}\in\{0,0.5,1.0,2.5,5.0\}$ on the same 50 shelf scenes, where $\lambda_{\mathrm{vis}}=0$ corresponds to Effort-Only.

As shown in Table~\ref{tab:sim_weight_sweep}, success remains constant as manipulation clears enough objects to eventually retrieve the target. Thus, allowing it is primarily responsible for recovering success under severe occlusion. Visibility prediction instead improves efficiency: without it, occluders are selected by arm motion alone, blind to whether their removal exposes the target, and from $\lambda_{\mathrm{vis}}=0$ to $2.5$, trajectory length decreases from $2.63$ to $2.37$, replans from $4.84$ to $4.50$, and removed occluders from $2.46$ to $2.12$. The metrics plateau beyond $\lambda_{\mathrm{vis}}=2.5$, indicating that additional visibility weight no longer changes the selected manipulation ordering. We therefore use $\lambda_{\mathrm{vis}}=2.5$ in the remaining experiments.

These results support H2: manipulation itself recovers task success, while imagined visibility makes that manipulation more selective and avoids unnecessary physical interaction.

\begin{table}[htbp]
\centering
\renewcommand{\arraystretch}{1.2}
\setlength{\tabcolsep}{4pt}
\resizebox{\columnwidth}{!}{%
\begin{tabular}{|l|c|c|c|c|}
\hline
\textbf{$\lambda_{\mathrm{vis}}$} & \textbf{Succ. \%} & \textbf{Traj.} $\downarrow$ & \textbf{Replans} $\downarrow$ & \textbf{Occl. Rem.} $\downarrow$ \\
\hline
0.0 (Effort-Only) & 84.0 & 2.63 & 4.84 & 2.46 \\
\hline
0.5 & 84.0 & 2.47 & 4.64 & 2.26 \\
\hline
1.0 & 84.0 & 2.39 & 4.52 & 2.14 \\
\hline
2.5 (default) & 84.0 & \textbf{2.37} & \textbf{4.50} & \textbf{2.12} \\
\hline
5.0 & 84.0 & \textbf{2.37} & \textbf{4.50} & \textbf{2.12} \\
\hline
\end{tabular}%
}
\caption{Visibility-weight sweep on the shelf ($n=50$ paired trials per weight; the effort weight $\lambda_{\mathrm{effort}}$ is held at $1.0$ throughout).}
\label{tab:sim_weight_sweep}
\vspace{-0.5cm}
\end{table}

\subsection{H3: Does Semantic Guidance Complement Geometry?}
\label{sec:exp-vlm}

We next test whether VLM-based belief shaping provides information complementary to geometric evaluation. We generate $n=30$ environments containing three spatially separated occluder clusters composed of cylinders, cubes, and pyramids. The hidden target matches the shape of exactly one cluster and is placed within that cluster. Both the target shape and the left-to-right ordering of the clusters are randomized independently across trials, preventing a fixed positional prior from solving the task.

The VLM receives the target color and shape together with a rendered image (Fig.~\ref{fig:vlm_input}), and predicts which cluster is most likely to contain the target. Its prediction softly reweights the particle belief rather than eliminating hypotheses. We additionally verify using the simulator segmentation mask that the hidden target contributes zero pixels to every VLM input, preventing direct visual detection of the target.

Table~\ref{tab:vlm_diagnostic} reports task success, VLM cluster accuracy, first-ghost accuracy, replans, and removed occluders. First-ghost accuracy measures whether the semantically reweighted belief places the determinized target hypothesis in the correct cluster; it can differ from VLM accuracy since semantic guidance is used as a soft prior and not a hard constraint.

Adding Gemini to UC raises success from $53.3\%$ to $73.3\%$ and reduces replans from $10.50$ to $6.27$, but unit-cost planning still fails 8 trials after exhausting the replan budget without removing an occluder. Thus, improving the belief alone cannot overcome the structural preference for short \textsc{Detect}-first skeletons. Conversely, MC without semantic guidance reaches $83.3\%$ success but requires $6.30$ replans and $3.90$ removals on average. Combining both components yields $100.0\%$ success with only $1.40$ replans and $0.40$ removals. Across VLMs, stronger belief guidance also generally reduces replanning and search effort. Semantic reasoning therefore determines \emph{where} to search, while geometric imagination determines \emph{how} to observe the resulting hypothesis.

\begin{figure}[ht]
    \centering
    \includegraphics[width=1.0\linewidth]{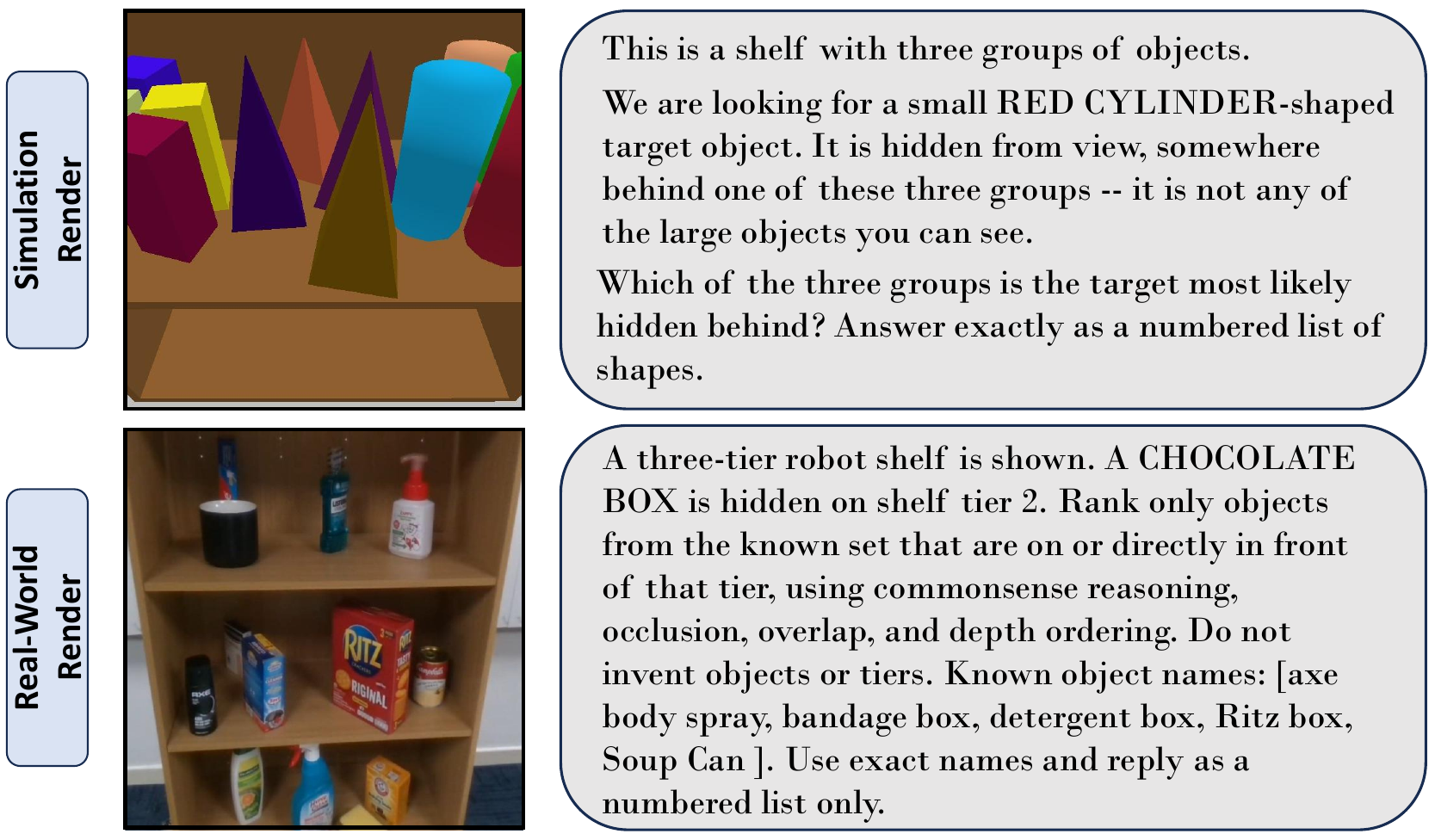}
        \caption{Example VLM input, PyBullet render(simulation) and RealSense capture(real-world) for the cluster scenes with corresponding prompts. }
    \label{fig:vlm_input}
    \label{fig:vlm_input}
    \vspace{-0.4cm}
\end{figure}

\begin{table}[htbp]
\centering
\renewcommand{\arraystretch}{1.2}
\setlength{\tabcolsep}{4pt}
\resizebox{\columnwidth}{!}{%
\begin{tabular}{|l|c|cc|c|c|}
\hline
\textbf{Method} & \textbf{Succ. \%} $\uparrow$ & \multicolumn{2}{c|}{\textbf{Accuracy \%} $\uparrow$} & \textbf{Replans} & \textbf{Occl.} \\
\cline{3-4}
& & \textbf{Cluster} & \textbf{1st-Ghost} & $\downarrow$ & \textbf{Rem.} $\downarrow$ \\
\hline
UC (no VLM) & 53.3 & -- & 40.0 & 10.50 & 0.00 \\
\hline
UC + Gemini 3.8 Flash & 73.3 & \textbf{100.0} & \textbf{93.3} & 6.27 & 0.00 \\
\hline
MC (no VLM) & 83.3 & -- & 40.0 & 6.30 & 3.90 \\
\hline
MC + MiniCPM-V & 83.3 & 53.3 & 53.3 & 4.87 & 2.67 \\
\hline
MC + \textsc{Qwen3-VL:32B} & 93.3 & 63.3 & 60.0 & 3.50 & 1.93 \\
\hline
MC + GPT-4o-mini & 93.3 & 86.7 & 76.7 & 2.63 & 1.17 \\
\hline
MC + Gemini 3.8 Flash & \textbf{100.0} & \textbf{100.0} & \textbf{93.3} & \textbf{1.40} & \textbf{0.40} \\
\hline
\end{tabular}%
}
\caption{VLM-guided belief biasing on cluster scenes ($n=30$ independent environments, one all-cylinder, one all-cube and one all-pyramid cluster each; target shape and slot order randomized per environment).}
\label{tab:vlm_diagnostic}
\vspace{-0.4cm}
\end{table}

\subsection{Real-World Validation}
\label{sec:real-world}

We validate the complete pipeline on real tabletop and shelf scenes containing everyday objects with partially or fully occluded targets; refer to Fig.~\ref{fig:fig_teaser} for a representative shelf trial. Unlike simulation, no complete scene model is provided in advance. A RealSense D415 captures the workspace; FoundationStereo~\cite{wen2025foundationstereo} estimates dense depth, SAM~3~\cite{carion2026sam} segments visible objects, SAM~3D~\cite{chen2026sam} reconstructs their geometry, and FoundationPose~\cite{wen2024foundationpose} estimates their 6-DoF poses. The resulting scene is updated whenever sensing or manipulation reveals previously occluded regions; unobserved geometry is not assumed to be known free space.

Semantic belief shaping uses the same RGB observation. The VLM only ranks visible objects according to how likely they are to occlude the target. For shelf scenes, the prompt restricts the ranking to objects on, or directly in front of, the relevant tier. This setting exercises real object semantics that are intentionally absent from the shape-based simulation study. For these experiments, we use a system with a dedicated NVIDIA RTX 5090 GPU, primarily to run the compute-heavy 3D perception models and the local VLM model. We use Qwen3-VL:32B as we found it sufficiently good at real-world semantic reasoning.

Compared with a geometry-only(no VLM) ablation using the same pipeline, \textsc{Imagine-TAMP} reduces mean replanning steps from $3.25$ to $1.50$ and total planning time from $232.0$\,s to $158.5$\,s, a $32\%$ reduction across 4 trials, while both variants complete the evaluated tasks. This study validates the complete system with real perception exhibiting the same reduction in physical search effort observed in simulation.

\section{Conclusion}

We present \textsc{Imagine-TAMP}, an imagination-guided planning and execution framework for PO-TAMP. Rather than replacing model-based planning, generative models provide inexpensive guidance for task-level decisions: a VLM shapes the belief over likely target locations, while geometric imagination estimates how sensing and manipulation affect target visibility. These estimates define non-unit costs for ranking symbolic plan skeletons before continuous refinement, and the selected strategy is still refined and verified by the underlying TAMP solver. Our experiments show that this improves sensing-versus-manipulation decisions under severe occlusion, while semantic belief shaping further reduces unnecessary search and replanning.


The current approach nevertheless reasons over limited geometric hypotheses and a finite set of candidate strategies, using approximate task-level costs. Future work will consider uncertainty over multiple imagined geometries and more adaptive generation and evaluation of task strategies.






\section{Acknowledgment}
The authors acknowledge the use of Gemini and ChatGPT to visually enhance some illustrations in Fig. 2 without altering their scientific content.

\bibliography{references}
\bibliographystyle{IEEEtran}

\end{document}